\documentclass{article}

\PassOptionsToPackage{numbers, compress}{natbib}
\usepackage[preprint]{neurips_2026}

\usepackage[utf8]{inputenc} 
\usepackage[T1]{fontenc}    
\usepackage{hyperref}       
\usepackage{url}            
\usepackage{booktabs}       
\usepackage{amsfonts}       
\usepackage{nicefrac}       
\usepackage{microtype}      
\usepackage{xcolor}         
\usepackage{amsmath, array, subcaption, graphicx}
\usepackage{float}

\title{Identifying and Mitigating Systemic Measurement Bias in Production LLM Inference Benchmarks}

\author{%
  Ashok Chandrasekar\\
  Google\\
  \texttt{achandrasekar@google.com}\\
  \And
  Jason Kramberger\\
  Google\\
  \texttt{jkramberger@google.com}\\
}

\begin{document}

\maketitle

\begin{abstract}
As Large Language Models (LLMs) transition from research environments to production deployments, evaluating their performance against strict Service Level Objectives (SLOs) has become critical. However, current evaluation methodologies suffer from severe measurement bias at scale. We demonstrate that widely used benchmarking utilities rely on single-process, asyncio-driven architectures that introduce fundamental client-side queuing bottlenecks under high concurrency. By modeling the benchmarking client as an $M/G/1$ queue, we mathematically demonstrate how the Python Global Interpreter Lock (GIL) artificially inflates Time to First Token (TTFT) and Time Per Output Token (TPOT) metrics as request rates scale. To resolve this systematic inaccuracy, we propose an unbiased, multi-process evaluation framework that effectively distributes client-side load, ensuring negligible queuing overhead. Furthermore, we formalize a composite metric, Normalized Time Per Output Token (NTPOT), to robustly amortize end-to-end latency, including prefill and scheduling delays across sequence lengths. Our empirical evaluation demonstrates that this methodology successfully isolates pure serving engine performance, enabling accurate, reproducible profiling of LLMs at production scales exceeding thousands of queries per second.
\end{abstract}

\section{\textbf{Introduction}}

The rapid adoption of LLMs has catalyzed a shift in how the AI community evaluates system performance. While early research heavily prioritized raw throughput, production systems are governed by strict user-facing SLOs, necessitating precise latency profiling. Consequently, the accuracy of the utilities used to benchmark serving frameworks such as vLLM \cite{kwon2023vllm}, SGLang \cite{zheng2023sglang}, NVIDIA TensorRT-LLM \cite{nvidia2023tensorrtllm}, and Hugging Face Text Generation Inference (TGI) \cite{hf2023tgi} directly dictates infrastructural decisions and architectural optimizations.

Despite the critical nature of these evaluations, we identify a systemic flaw in standard benchmarking methodologies: the tools used to measure high-concurrency systems are fundamentally bottlenecked by their own architectures. Existing benchmarking utilities \cite{vllm2023benchmarks}, \cite{semianalysis2025inferencex}, \cite{nvidia2024genaiperf} typically employ single-process Python scripts relying on asynchronous event loops. While sufficient for low-load testing, we empirically demonstrate that under production-scale request rates (100s to 1000s of Queries Per Second), the Python Global Interpreter Lock (GIL) fundamentally throttles the client's service rate.

This limitation introduces a severe measurement bias into LLM evaluation. When utilizing a saturated, single-process client to evaluate a high-concurrency server, the benchmarking utility artificially inflates the resulting latency metrics. By modeling the evaluation client as an $M/G/1$ queue, we show that as the utilization factor of the event loop approaches saturation, client-side wait times approach infinity. Because standard benchmarks record latency upon processing a token rather than its network arrival, this client-side bottleneck is erroneously recorded as model server degradation. Consequently, what is frequently diagnosed as an infrastructural failure is, in reality, a measurement bias induced by the observer itself.

Furthermore, traditional latency metrics fail to accurately capture the true computational cost of serving. The emergence of prefix caching has rendered initial response times highly variable, while standard TPOT measurements obscure heavy prefill computations by focusing solely on the decode phase.

To address these methodological shortcomings, this paper introduces a rigorous framework for unbiased LLM evaluation. Our specific contributions are threefold:

\begin{enumerate}
    \item \textbf{Empirical Isolation of Measurement Bias:} We provide a quantitative analysis of client-side queuing bottlenecks, demonstrating mathematically and empirically how single-process architectures artificially inflate latency metrics under high concurrency.
    \item \textbf{Formalization of a Unified SLO Metric:} We define Normalized Time Per Output Token (NTPOT), a robust composite metric that amortizes end-to-end request latency, incorporating both queue contention and prefill penalties across output sequences to standardize performance evaluation.
    \item  \textbf{Design of an Unbiased Evaluation Framework:} We propose a multi-process benchmarking methodology that mitigates client-side measurement bias. By integrating NTPOT with continuous latency profiling, this framework enables the empirical discovery of true hardware saturation points at production scale.
\end{enumerate}

\section{\textbf{Background and Related Work}}

The deployment of LLM inference workloads in production environments is highly complex, necessitating the execution of models across diverse accelerator and machine architectures. These environments frequently manage heterogeneous workloads, such as serving various LoRA adapters \cite{hu2022lora} or concurrently serving models of different sizes, while accommodating fluctuating traffic patterns, including diurnal variations and severe traffic spikes \cite{jaiswal2025sageserve}. Modern serving strategies further complicate evaluation by employing prefill/decode disaggregation \cite{zhong2024distserve}, \cite{patel2024splitwise} and various parallelism modalities, including tensor parallelism (TP) \cite{shoeybi2019megatron}, pipeline parallelism (PP) \cite{narayanan2021efficient}, data parallelism (DP) \cite{li2020pytorch}, and expert parallelism (EP) \cite{lepikhin2020gshard} for Mixture-of-Experts (MoE) models \cite{shazeer2017outrageously}.

Given these architectural complexities, achieving an unbiased, production-grade benchmark requires satisfying strict methodological prerequisites:

\begin{enumerate}
    \item \textbf{High-Concurrency Load Generation:} The methodology must sustain high queries per second (QPS) to accurately map the throughput vs. latency curve and identify true hardware saturation thresholds.
    \item \textbf{Representative Workload Simulation:} Evaluations must utilize real-world prompt datasets and simulate complex traffic shaping (e.g., burst traffic, traffic splitting) rather than relying on uniform, randomly generated tokens.
    \item \textbf{Granular LLM Metrics:} The framework must capture LLM-specific token-level latency metrics (TTFT, TPOT, ITL) while remaining independent of the underlying server backend.
\end{enumerate}

Despite these clear prerequisites, an analysis of the current benchmarking landscape reveals significant methodological fragmentation. Standardized hardware performance evaluation frameworks like MLPerf \cite{reddi2020mlperf} or analyst utilities like Semi Analysis' Inference X \cite{semianalysis2025inferencex} evaluate systems as "black box" endpoints with the primary objective of demonstrating competitive accelerator performance \cite{mlcommons2024v4}, \cite{semianalysis2025comp}. Other analysts like Artificial Analysis \cite{artificialanalysis2025} or the now-deprecated LLM Perf \cite{rayproject2025llmperf} prioritize the benchmarking of hosted inference offerings like Google's Vertex, Amazon Bedrock, etc. In general, they lack the granular, high-load stress testing capabilities necessary for evaluating a production serving stack. Conversely, general-purpose HTTP load testers (e.g., K6 \cite{grafana2021k6}, and Locust \cite{heyman2011locust}) successfully manage high concurrency but are fundamentally decoupled from the specific streaming transport semantics and unbounded output domains of generative AI.

To bridge this gap, an industry-wide anti-pattern has emerged where practitioners misapply model server micro-benchmarking utilities (e.g., vLLM Bench \cite{vllm2023benchmarks}, SGLang benchmark \cite{sglangbenchmarks}, NVIDIA GenAI Perf \cite{nvidia2024genaiperf}) for production-level evaluations. Because these tools rely heavily on single-process Python \textit{asyncio} architectures, adapting them for high-load concurrency introduces fundamental client-side constraints that skew the resulting data. To resolve these systemic constraints, a new class of multi-process utilities like GuideLLM \cite{neuralmagic2024guidellm} and NVIDIA AI Perf \cite{nvidia2024aiperf} has recently emerged. However, the theoretical foundations validating this architectural shift remain unformalized, and as our empirical evaluation reveals, early implementations of this paradigm still encounter structural bottlenecks at high concurrency. In this paper, we formalize this distributed evaluation methodology and concretely instantiate it through \textbf{Inference Perf} \cite{inferenceperf2026}, an open-source evaluation system engineered to optimally operationalize this architecture. Through this implementation, we demonstrate how multi-process load partitioning mathematically mitigates client-side measurement bias, while empirically proving that Inference Perf sustains significantly higher load fidelity than contemporary multi-process alternatives at production scales exceeding 5,000 QPS.

\section{\textbf{Continuous Latency Profiling and Saturation Detection}}

The evaluation of LLM inference systems cannot be reduced to static, single-point measurements. As established in our analysis of evaluation methodologies, measuring a serving stack at an arbitrary, static concurrency limit obscures the complex queuing dynamics that emerge under dynamic production traffic. To accurately characterize the true operational capacity of an inference engine, performance must be empirically mapped across a continuous load spectrum.

To achieve this, we advocate for the standardized use of \textbf{Latency Profiles}: empirical mappings of a system's latency versus throughput frontier. A latency profile is generated by executing an unbiased benchmark that initiates at a baseline arrival rate and systematically sweeps the load via linear or geometric progression until the serving stack reaches full structural saturation. See Appendix A.1 for a comprehensive, empirical latency profile generated by Inference Perf.

\subsection{Saturation Detection and Latency Normalization}
The primary utility of a latency profile is the precise identification of the server's saturation inflection point, the critical threshold where the system achieves maximum token throughput immediately prior to queuing degradation. Optimizing hardware utilization inherently requires operating the serving stack right below this threshold to avoid the exponential latency penalties associated with queue over-utilization. However, mapping standard end-to-end latency on the y-axis introduces a critical methodological flaw: request latency scales linearly with input and output sequence lengths, making it impossible to directly compare profiles across disparate models or datasets. To construct a universal performance curve, we must normalize the latency measurement against the actual computational work performed by the engine.

\subsection{Formalization and Advantages of NTPOT}
We propose the use of Normalized Time per Output Token (NTPOT) for latency reporting. For a given inference request $i$, we define $L_i$ as the total end-to-end latency and $N_i$ as the total number of generated output tokens:$$ \text{NTPOT}_i = \frac{L_i}{N_i} $$To capture the comprehensive computational expense in proportion to its magnitude, we expand this definition to isolate the underlying serving phases:$$ \text{NTPOT}_i = \frac{T_{\text{queue}}^{(i)} + (I_i \times T_{\text{prefill}}^{(i)}) + (N_i \times T_{\text{decode}}^{(i)})}{N_i} $$Where $T_{\text{queue}}^{(i)}$ is the server-side wait time, $I_i$ is the input length, $T_{\text{prefill}}^{(i)}$ is the prefill token latency, and $T_{\text{decode}}^{(i)}$ is the decode token latency. In a production scenario, individual measurements are subject to system jitter. Therefore, for a benchmark comprising $M$ requests, performance must be evaluated across the mean $\mathbb{E}[\text{NTPOT}]$ and sample variance to quantify stability. Unlike isolated decode metrics, NTPOT is a composite metric of inference latencies. In highly loaded systems, pure decode TPOT often remains relatively flat, while queue times and prefill compute degrade exponentially. By encapsulating these initial delays and dividing them by the output length, a spike in p99 $\text{NTPOT}$ immediately alerts teams that scheduling capacity is saturated. Furthermore, it intrinsically normalizes the massive prefill costs associated with Retrieval-Augmented Generation (RAG) workflows, accurately reflecting the wait time the end user experiences before receiving the payload.

\section{\textbf{Confounding Variables and Reproducibility Anti-Patterns}}

To establish a reproducible evaluation standard, we must systematically isolate the variables that introduce measurement bias during the request lifecycle. A standard Python-based evaluation client initiates asynchronous tasks to manage a target QPS, awaits token delivery via streaming requests, and subsequently calculates metrics such as TTFT and TPOT. However, several systemic anti-patterns frequently impact this process.

\subsection{\textbf{Client-Side Utilization Constraints}}

Most common utilities, including vLLM, SGLang, and competitive benchmarks like Semi Analysis's Inference X, utilize single-process Python scripts. These rely on asyncio to manage concurrent requests. However, the load capacity of such tools is fundamentally constrained by the single CPU core and its clock speed because of Python’s Global Interpreter Lock (GIL), which allows only one thread to execute at a time.  

To understand why this architecture fundamentally skews performance metrics at scale, we model the benchmarking client's single-threaded event loop as an $M/G/1$ queue.

Let $\lambda$ represent the arrival rate of incoming server responses (tokens) across all concurrent streams. Let $\mu$ represent the service rate, defined as the speed at which the Python event loop can deserialize and record a single token. The utilization factor of the event loop is therefore $\rho = \frac{\lambda}{\mu}$.

According to the Pollaczek-Khinchine formula \cite{kleinrock1975queueing}, the expected waiting time in the client's internal processing queue, $W_q$, is:$$ W_q = \frac{\lambda \mathbb{E}[S^2]}{2(1 - \rho)} $$Where $\mathbb{E}[S^2]$ is the second moment of the client's service time distribution. As the user-requested QPS increases toward production loads of 1,000 or 5,000, the aggregate token arrival rate $\lambda$ rapidly approaches the event loop's maximum service capacity $\mu$. As $\rho \to 1$, the client-side wait time $W_q \to \infty$.

In production evaluations, this queuing bottleneck on the client side leads to two primary issues:

\begin{enumerate}
    \item \textbf{Rate Failure}: The tool may fail to hit the requested QPS without providing explicit warnings. This is often only detectable by analyzing actual run durations against the expected rate.

    \item \textbf{Latency Inflation}: Uncapped asyncio tasks cause significant slowdowns due to context switching overhead, resulting in inaccurately high reported latency numbers. Because the benchmark records latency upon processing the token rather than its network arrival, the measured TTFT and TPOT are heavily inflated by $W_q$.
\end{enumerate}

By distributing $\lambda$ across $k$ isolated Python processes, a multi-process architecture (such as Inference Perf) reduces the utilization factor per process to $\rho_k = \frac{\lambda}{k \mu}$. This structural change ensures that $\rho_k$ remains well below 1, keeping $W_q$ negligible and ensuring that reported latency metrics accurately reflect the server's performance rather than the client's bottleneck.

\subsection{\textbf{The Illusion of Concurrency-Based Load Generation}}

Due to the rate failures inherent in achieving a target QPS on single-process architectures, many modern benchmarking utilities have deprecated strict QPS targeting in favor of concurrency-based load definition. While defining load via a fixed number of concurrent threads or asyncio tasks simplifies execution on restricted hardware, it fundamentally obscures system degradation. Concurrency bounds do not resolve the $M/G/1$ queuing bottleneck; they merely cap the arrival rate, $\lambda$, to artificially prevent the event loop's utilization factor, $\rho$, from approaching or exceeding $1$. Consequently, evaluations relying strictly on bounded concurrency fail to expose the true latency spikes associated with the unthrottled, dynamic traffic patterns characteristic of production deployments.

\subsection{\textbf{Impact of Stochastic Sampling Parameters on Throughput}}

Benchmarking reproducibility is severely compromised when stochastic sampling parameters are inconsistently applied. For instance, modifying the model's temperature parameter fundamentally alters the decoding phase's arithmetic intensity \cite{williams2009roofline},  \cite{chen2022transformer}. Setting the temperature to $0$ disables probabilistic sampling, forcing the model into Greedy Scheduling, which eliminates the need to evaluate lower-probability branches and artificially accelerates token generation. As detailed in Appendix A.2, identical hardware processing 1,000 requests exhibited a 21\% throughput variance solely due to temperature configuration (6,767 tok/s at Temp=0 versus 5,573 tok/s at Temp=0.7). Benchmarks that default to a temperature of 0 fail to reflect the compute requirements of real-world, non-deterministic generative workloads.

\subsection{\textbf{Context-Awareness and Prefix Caching Bias}}

Discrepancies in performance results frequently arise between different benchmarking tools even when utilizing identical public datasets, such as HuggingFace ShareGPT. These variations are often artifacts of how the client handles prompt truncation, random sampling, and request sequencing. As illustrated in Appendix A.2, two different utilities processing the exact same dataset generated a 50\% discrepancy in the volume of processed input tokens. Furthermore, server-side optimizations like prefix caching introduce extreme variance into TTFT measurements based purely on request sequencing. To ensure empirical validity, benchmarking frameworks must explicitly report the enforced cache hit ratio and strictly standardize truncation limits across all evaluation runs.

\section{\textbf{Empirical Evaluation of Measurement Bias at Scale}}

This section presents a comparative empirical analysis to quantify the measurement bias introduced by single-process benchmarking architectures under high-concurrency conditions. To isolate client-side evaluation overhead from actual model server performance, we conducted testing against the \textit{llm-d-inference-sim} simulator \cite{llmd2026inference}. This simulator models production server traffic and returns tokens without added computational delay, guaranteeing that any observed performance degradation is strictly an artifact of the benchmarking utility.

To ensure controlled experimentation, all evaluations utilized fixed-length random data generation, configured strictly to 320 input tokens and 200 output tokens. The benchmarking client was hosted on an identically configured c4-standard-144 instance, utilizing a 6th generation Intel Xeon processor with 144 vCPUs, to eliminate hardware-induced bottlenecks. The methodological paradigms evaluated in this study are detailed in Table 1.

\begin{table}
\centering
\caption{Classification of LLM benchmarking tools}
\label{tab:1}
\begin{tabular}{ l  l  l }\toprule

\textbf{Utility} & \textbf{Architecture} & \textbf{Primary Application} \\\midrule

vLLM Bench & Single-Process & Micro Benchmarking \\

Semi Analysis Inference X & Single-Process & Competitive Black-box API Evaluation\\

Inference Perf & Multi-Process & Production Scale Evaluation \\

GuideLLM & Multi-Process & Production Scale Evaluation \\

NVIDIA AI Perf & Multi-Process & Production Scale Evaluation \\ \bottomrule

\end{tabular}

\end{table}

\begin{table}
\centering
\caption{Effect of hardware and single vs. multi-process architectures on QPS and throughput}
\label{tab:2}
\begin{tabular}{>{\raggedright\arraybackslash}p{0.17\linewidth}>{\raggedright\arraybackslash}p{0.3\linewidth}>{\raggedright\arraybackslash}p{0.17\linewidth}>{\raggedright\arraybackslash}p{0.17\linewidth}>{\raggedright\arraybackslash}p{0.19\linewidth}}
\toprule
\textbf{Tool} & \textbf{Benchmarking Hardware} & \textbf{User Requested QPS} & \textbf{Actual Request QPS} & \textbf{Total Token Throughput (tok/s)} \\
\midrule
Single-process& E2-medium (2 vCPU shared core Intel) & 100 & 38.59 & 19410 \\

Multi-process& E2-medium (2 vCPU shared core Intel) & 100 & 65.47 & 31702 \\

Single-process & C4-standard-144 (144 vCPU Intel 6th Gen Xeon)& 200 & 171.47 & 86258 \\

Multi-process& C4-standard-144 (144 vCPU Intel 6th Gen Xeon)& 200 & 199.57 & 96623 \\
\bottomrule

\end{tabular}

\end{table}

\subsection{\textbf{Hardware Constraints on Load Generation}}

Before analyzing the methodological biases at extreme concurrency, we must first isolate the impact of the underlying client-side hardware on load generation capacity. Client hardware specifications significantly influence benchmarking load fidelity. For example, our baseline testing revealed that a single-process architecture executing on an entry-level e2-medium instance saturated at merely 38.59 QPS. Conversely, executing our multi-process methodology on a 144-vCPU instance successfully achieved the target 200 QPS baseline without throttling (see Table 2 for the full hardware ablation study with Inference X as the single-process tool and Inference Perf as the multi-process tool).

If the benchmarking tool itself acts as the bottleneck, the reported throughput becomes deceptive. Consequently, to mathematically eliminate hardware-induced bottlenecks from our analysis, all subsequent high-concurrency trials (1,000 and 5,000 QPS) were conducted strictly on the c4-standard-144 instance.

\begin{figure}[t]
  \centering
  \begin{subfigure}{.50\columnwidth}
    \centering
    \includegraphics[width=\linewidth]{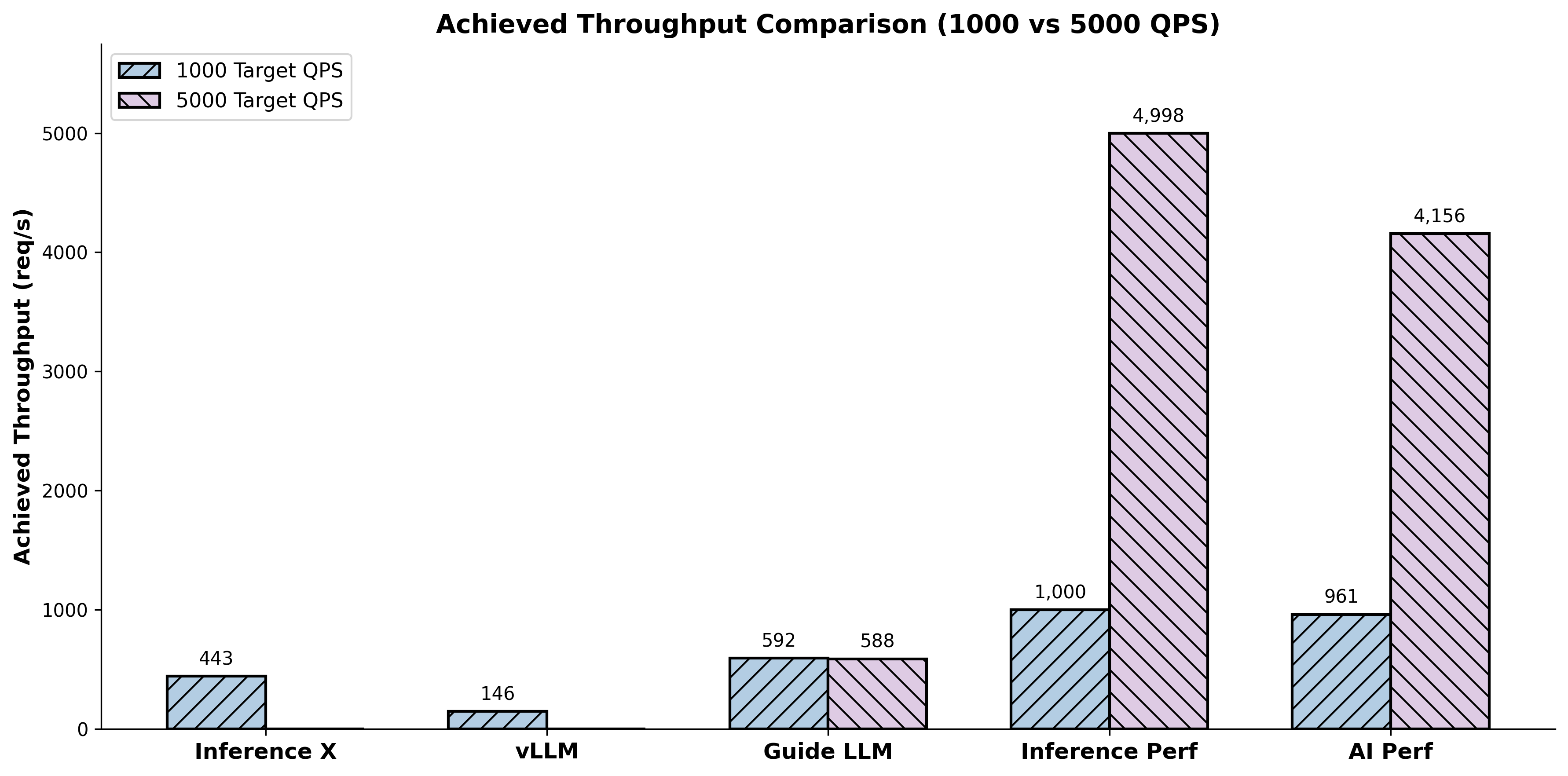}
    \caption{Throughput in QPS}
  \end{subfigure}%
  \hfill
  \begin{subfigure}{.50\columnwidth}
    \centering
    \includegraphics[width=\linewidth]{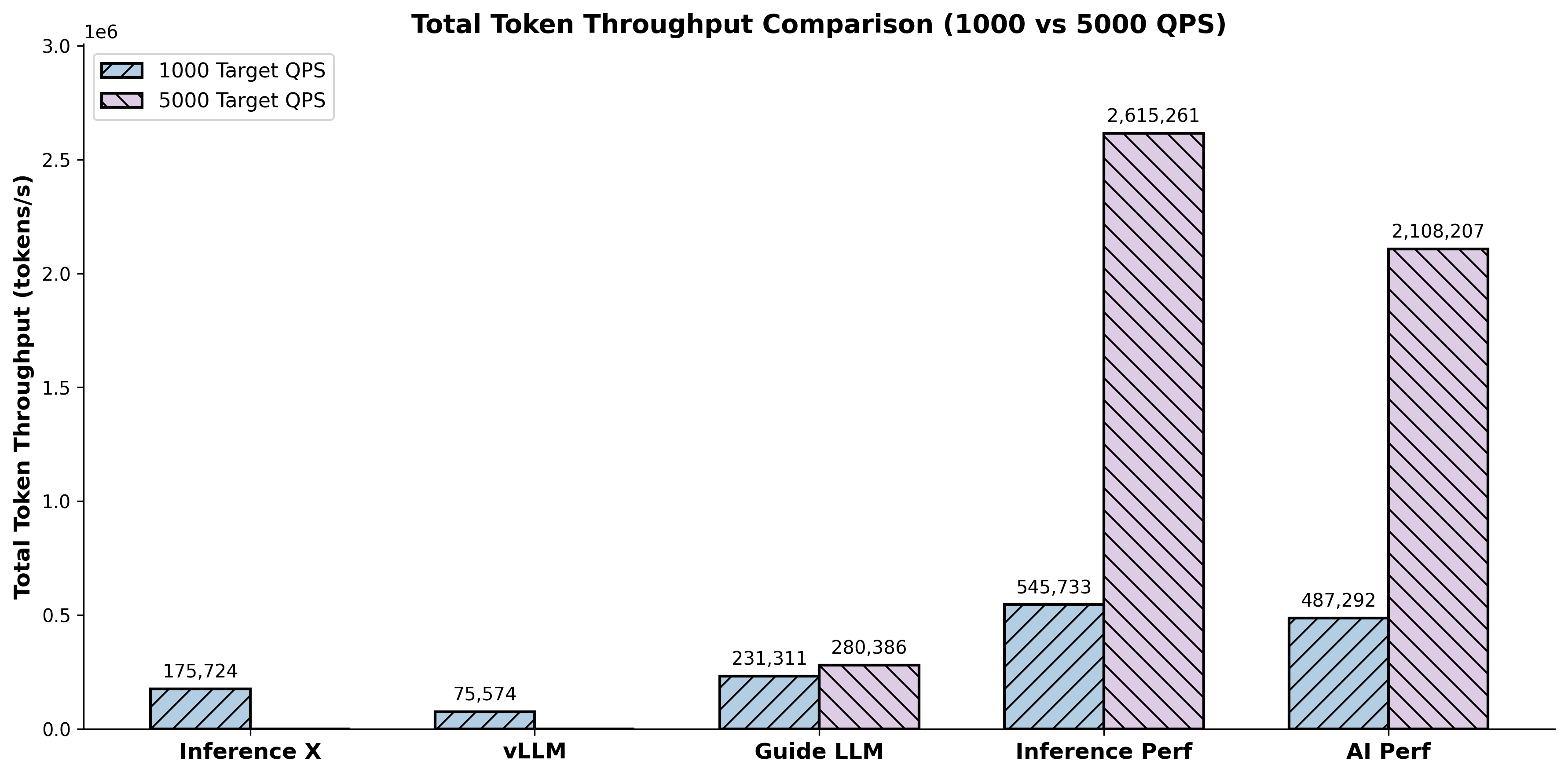}
    \caption{Throughput in tokens/second}
  \end{subfigure}%
  \hfill
  \begin{subfigure}{.50\columnwidth}
    \centering
    \includegraphics[width=\linewidth]{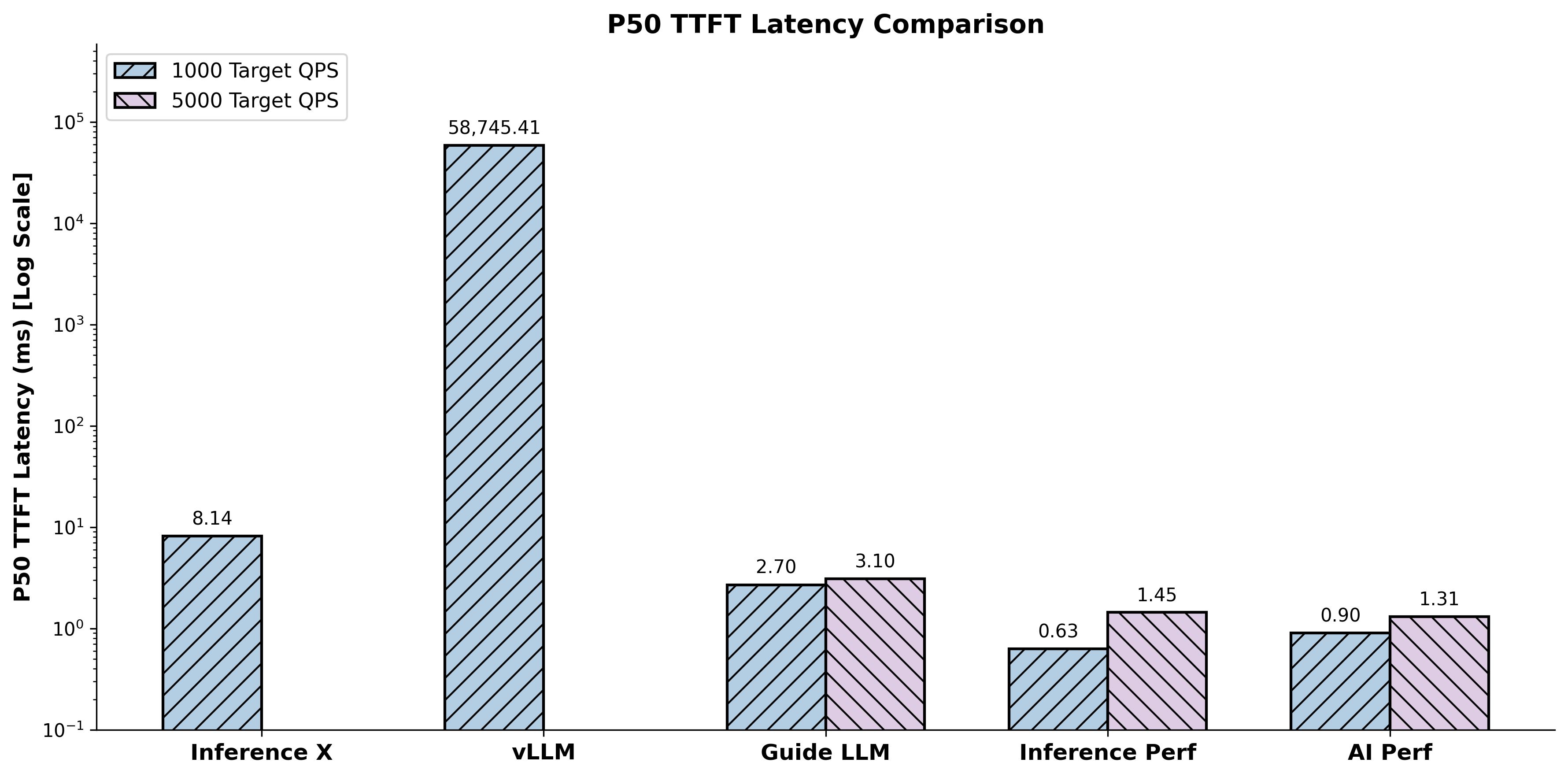}
    \caption{Median TTFT latency in ms}
  \end{subfigure}
  \caption{Analysis of throughput and latency. (a) and (b) demonstrate the rate failure of single-process tools, while (c) highlights the exponential inflation of TTFT}
\end{figure}

\subsection{\textbf{Analysis of Load Fidelity and Rate Failure}}

The data in Figure 1a validates our theoretical model regarding the limitations of single-threaded \textit{asyncio} event loops. When tasked with target arrival rates of 1,000 and 5,000 QPS, single-process utilities exhibited catastrophic rate failure. At the 1,000 QPS threshold, vLLM Bench and Inference X saturated at 146 and 443 QPS, respectively. Conversely, multi-process frameworks maintained higher load fidelity, with Inference Perf successfully sustaining the requested 1,000 QPS load without throttling.

At extreme concurrency (5,000 QPS), the utilization factor $\rho$ of single-process tools exceeded 1, causing vLLM Bench and Inference X to experience timeouts and excessive error rates, necessitating their exclusion from the final results. The distributed multi-process methodology of Inference Perf successfully processed 4,998 QPS, demonstrating that partitioning the load generation across isolated processes effectively mitigates the event loop bottleneck.

Figure 1b illustrates the total token throughput (input and output) processed during the trials. The performance degradation of single-process architectures is stark: at 1,000 QPS, vLLM Bench processed only 75,574 tokens compared to the 545,733 tokens processed by the distributed architecture, representing a 7.2x discrepancy in measurement capacity. At 5,000 QPS, Inference Perf processed 2.6 million tokens, outperforming the nearest alternative multi-process tool (AI Perf) by 24\%.

\subsection{\textbf{Quantification of Latency Inflation}}

The most critical evidence of measurement bias is observed in the latency discrepancies illustrated in Figure 1c. Because the simulation environment imposes near-zero server-side latency, any recorded latency is purely client-side evaluation overhead.

During the 1,000 QPS trial, single-process utilities injected massive artificial delays, adding overhead ranging from 8 milliseconds to an extreme 58 seconds. In stark contrast, the optimized multi-process architecture of Inference Perf maintained an end-to-end overhead of merely 0.63 milliseconds.

This massive disparity confirms our hypothesis: as the single CPU core managing the event loop becomes overwhelmed by context switching and concurrent request dispatch, it fails to record token arrivals accurately. Consequently, single-process tools severely inflate TTFT and end-to-end latency metrics at scale. Relying on such tools for production evaluation will inevitably lead practitioners to diagnose non-existent server bottlenecks, proving the necessity of distributed evaluation frameworks for mathematically sound benchmarking.

\begin{figure}[t]
  \centering
  \begin{subfigure}{.48\columnwidth}
    \centering
    \includegraphics[width=\linewidth]{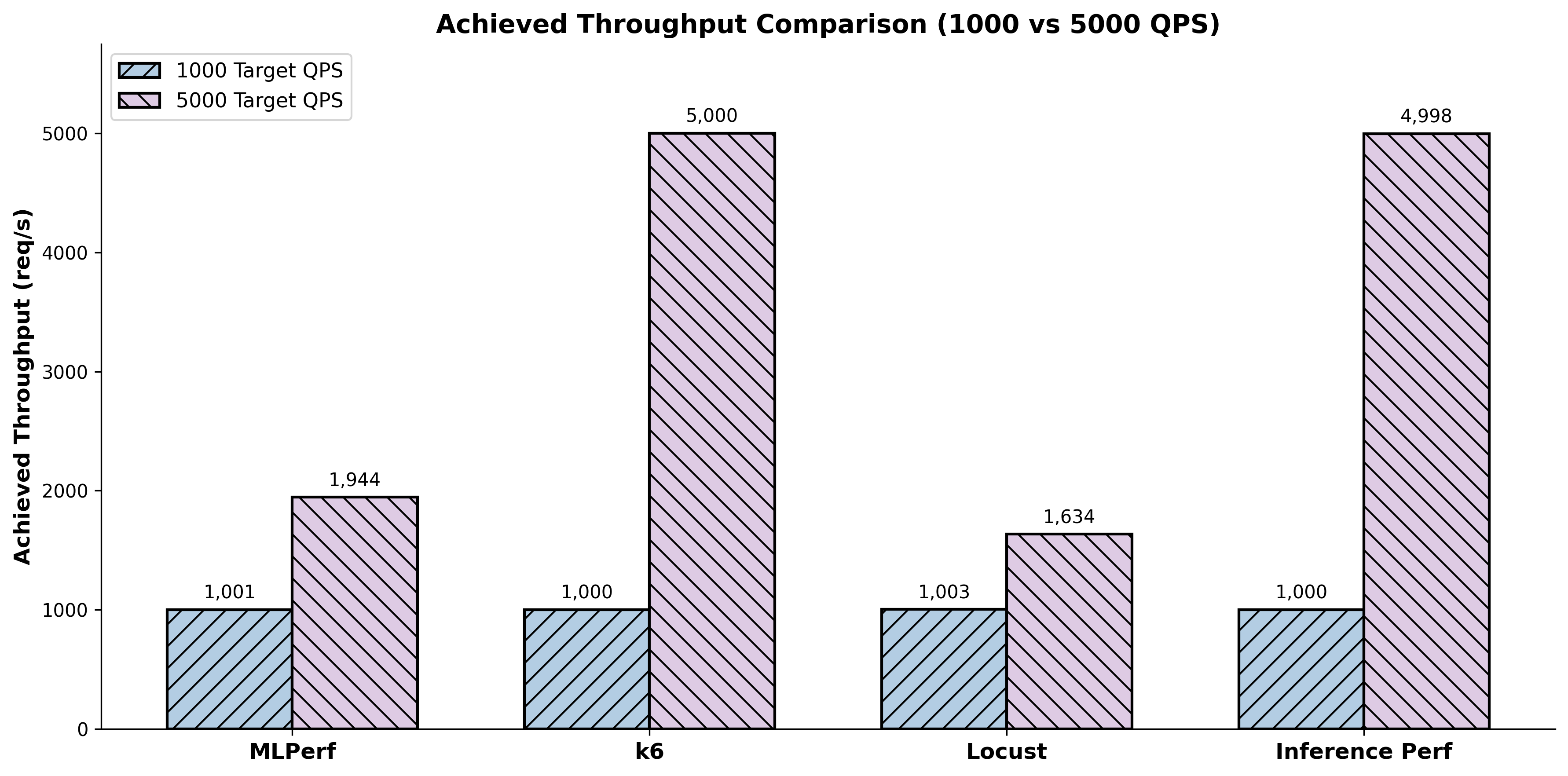}
    \caption{Throughput in QPS}
  \end{subfigure}%
  \hfill
  \begin{subfigure}{.48\columnwidth}
    \centering
    \includegraphics[width=\linewidth]{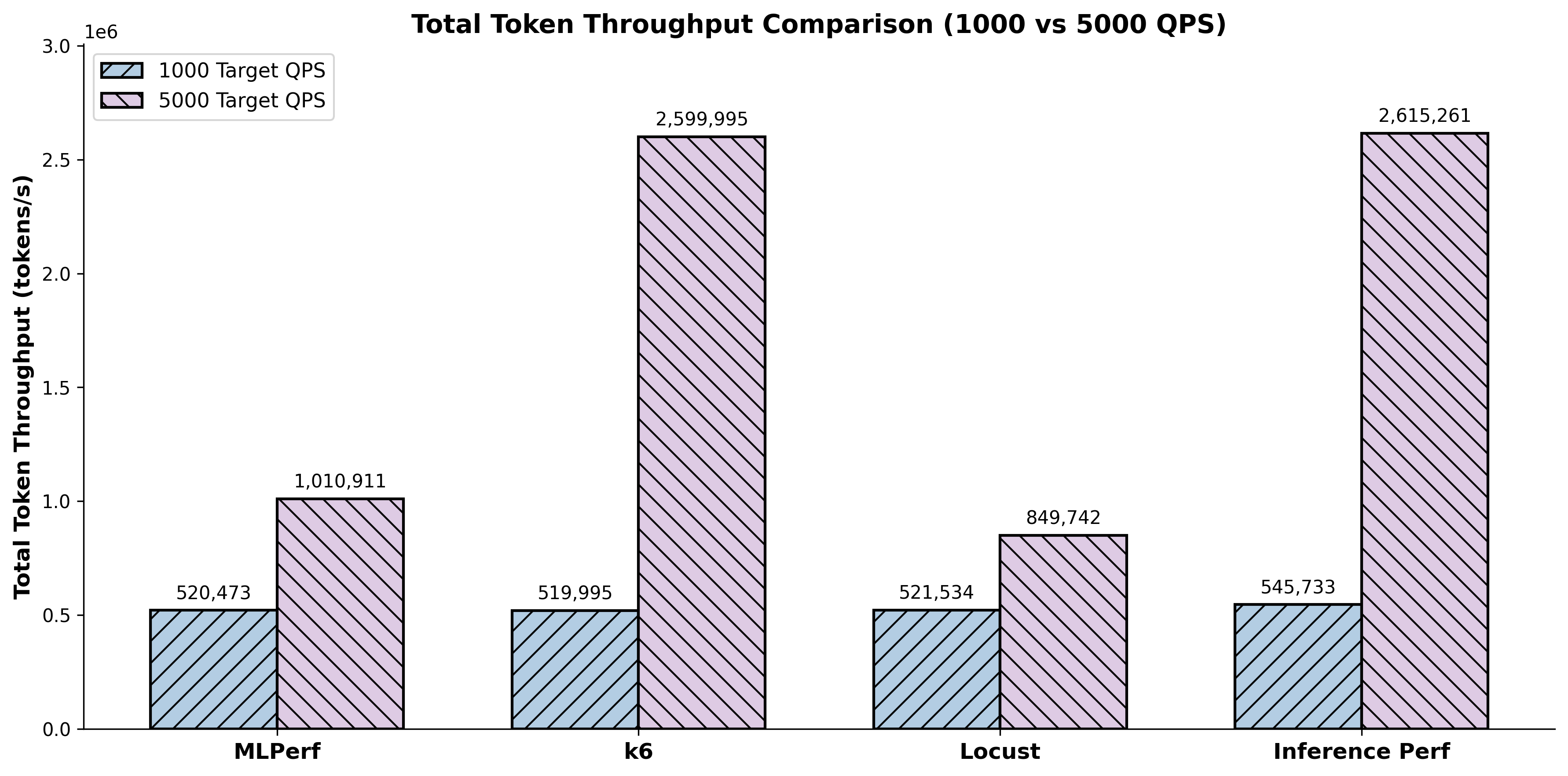}
    \caption{Throughput in tokens/second}
  \end{subfigure}%
  \caption{Analysis of throughput and latency against native HTTP load generators}
\end{figure}

\subsection{\textbf{Cross-Paradigm Validation Against Native HTTP Generators}}

To definitively isolate the Python Global Interpreter Lock (GIL) as the primary constraint on high-concurrency evaluation, we conducted a cross-paradigm validation against established, high-scale HTTP load generators. This comparison evaluated the proposed multi-process methodology against Grafana k6 (utilizing Go's native goroutines for concurrency), Locust (a Python-based distributed tester restricted here to a single node), and MLPerf's load generator (featuring a C++ core interfacing with Python bindings).

Because general-purpose HTTP tools lack native support for continuous token-level latency calculations, we restricted this evaluation to non-streaming standard HTTP requests. This methodological constraint theoretically favored the baseline utilities by eliminating the heavy deserialization overhead required for LLM token metrics.

While all utilities successfully sustained the 1,000 QPS baseline, extreme concurrency testing at 5,000 QPS revealed critical architectural limitations (see Figure 2). Locust and MLPerf reached saturation ceilings at 1,634 and 1,944 QPS, respectively. Despite MLPerf's C++ execution core, its reliance on a Python-based request processing interface reintroduced the GIL bottleneck under high load, causing the event loop utilization factor $\rho$ to hit saturation. Similarly, Locust's Python architecture constrained its theoretical capacity despite executing on a 144-vCPU instance.

In contrast, the Go-based k6 framework successfully bypassed these constraints via native multi-threading, sustaining the full 5,000 QPS load. Crucially, the proposed multi-process Python methodology achieved complete performance parity with k6, successfully processing the 5,000 QPS target without rate failure.

These results yield a vital architectural conclusion: while Python's single-threaded nature historically disqualifies it from extreme-scale benchmarking, its rich ecosystem of LLM libraries and tokenizers can be safely leveraged for production evaluation \textit{if and only if} the underlying execution relies on a distributed, multi-process architecture to bypass the GIL.

\section{\textbf{Design of an Unbiased Evaluation Framework }}

To address the aforementioned benchmarking challenges and eliminate systemic measurement bias, we propose a distributed evaluation methodology. We utilize Inference Perf as an experimental apparatus to demonstrate how this framework effectively isolates pure serving engine performance.

\subsection{\textbf{Distributed Load Partitioning}}

To bypass the client-side queuing bottlenecks inherent in single-process architectures, the framework employs a distributed multi-process design. The evaluation load is partitioned by spawning $m$ isolated processes corresponding to the available vCPUs on the host machine. Within these processes, Python’s \textit{asyncio} coroutines manage streaming server requests, allowing each worker to execute up to $n$ concurrent coroutines dictated by queue demand.

During execution, a data generator produces prompts from real or synthetic datasets based on specified distribution constraints. A load generator sequentially queues these requests according to user-defined arrival patterns (e.g., Poisson or constant distributions). Available workers pull from this queue to initiate tasks and track individual token arrival times. By distributing the load, this architecture mitigates single-threaded event loop saturation, ensuring the aggregated data accurately reflects server performance (refer to Appendix A.3 for a complete architectural flow diagram).

\subsection{\textbf{Dynamic Concurrency Regulation}}

To maintain strict experimental control, researchers can exercise precise regulation over the architecture via \textit{max\_workers} and \textit{worker\_max\_concurrency} parameters. While manual tuning is permitted, the framework is mathematically designed to automatically determine the optimal values based on the target load and available hardware resources. This ensures the client continuously operates below its utilization threshold and does not inadvertently throttle the evaluation.

\subsection{\textbf{Empirical Validation of Load Fidelity}}

To guarantee the integrity of the benchmark, the framework provides continuous observability into its own operational efficiency, ensuring that hardware limitations or external interference do not skew the data. We introduce two critical metrics to quantify measurement fidelity:

\begin{itemize}
    \item \textbf{Scheduling Precision:} The framework tracks the distribution of delays between a request’s intended dispatch time and its actual transmission. For results to be scientifically valid, the median, P90, and P99 of these delays must remain under one millisecond, ensuring target QPS is achieved without artificially injected latency.
    \item \textbf{Dispatch Throughput:} Unlike traditional tools that strictly report response-based QPS, this methodology actively monitors the actual QPS successfully transmitted by the load generator. This guarantees the evaluation adhered to the requested load pattern.
\end{itemize}

\subsection{\textbf{Methodological Reproducibility and Constraints}}

Scientific rigor demands exact reproducibility. The framework utilizes a declarative configuration system that governs request handling, dataset selection, and load generation. By archiving both user-defined parameters and default settings within a report directory, external researchers can identically reproduce the benchmark environment and prevent unintended consequences from variables like API temperature.

While this methodology reliably supports loads exceeding 10,000 QPS, it currently relies on vertical scaling across multi-core hardware. Scaling to extreme thresholds (e.g., 200,000+ QPS) may eventually necessitate a multi-node load-generation architecture. Furthermore, while Python introduces specific concurrency constraints, its unparalleled ecosystem of LLM libraries and tokenizers makes it the optimal foundation for reproducible, community-driven evaluation tooling.

\section{\textbf{Limitations and Future Work}}

While this study establishes a robust methodological foundation for unbiased LLM evaluation, several avenues for future research remain critical to capturing the full complexity of production inference environments.

\subsection{Multi-Node Distributed Evaluation}

Our empirical validation demonstrates that a localized, multi-process architecture successfully maintains load fidelity up to 5,000 QPS. However, as frontier models are increasingly deployed across massive, multi-datacenter clusters, evaluating extreme thresholds (e.g., >100,000 QPS) will eventually exceed the physical vCPU and network limitations of any single load-generating node. Future work must explore the theoretical and practical constraints of coordinating distributed, multi-node benchmarking frameworks, specifically addressing clock synchronization drift across nodes and network topology impacts on latency distributions.

\subsection{Mathematical Isolation of Prefix Caching}

As highlighted in our analysis of the prefill penalty, prefix caching introduces high variance into initial response times. While NTPOT effectively amortizes this variance across the output sequence, future research should develop standardized, mathematically constrained prompt datasets designed to guarantee exact cache-hit ratios (e.g., strict $0\%$, $50\%$, and $100\%$ shared prefix lengths). Formalizing these datasets will allow researchers to isolate the computational efficiency of a serving engine's radix tree or block-matching algorithms independent of raw, un-cached compute speed. 

\subsection{Modeling Non-Uniform Traffic Distributions}

The current framework assumes standard mathematical arrival distributions, such as constant or Poisson processes. However, real-world LLM traffic is frequently highly bursty and non-uniform. Future methodological enhancements should incorporate heavy-tailed Pareto distributions and Markov-modulated Poisson processes to simulate the "thundering herd" traffic spikes common in consumer-facing chatbots, allowing researchers to evaluate the resilience of continuous batching schedulers under sudden degradation.

\section{\textbf{Conclusion}}

As the AI community transitions from focusing purely on model capabilities to the realities of production deployment, the scientific rigor of our evaluation methodologies must scale commensurately. This paper identifies a critical, systemic measurement bias in the current LLM benchmarking ecosystem: the pervasive reliance on single-process \textit{asyncio} architectures for high-concurrency evaluation.

By modeling the benchmarking client as an $M/G/1$ queue, we demonstrated both theoretically and empirically that these architectures become fundamentally bottlenecked at scale. As utilization reaches saturation, client-side wait times artificially inflate reported metrics, leading practitioners to misdiagnose measurement artifacts as model server failures. To resolve this, we introduced a distributed, multi-process evaluation framework that effectively partitions load generation, maintaining absolute scheduling precision and dispatch throughput even at 5,000 QPS.

Furthermore, we formalized Normalized Time Per Output Token to address the inadequacies of standard latency metrics. By amortizing the entirety of the request lifecycle, including queue contention and heavy prefill computations across the output sequence, NTPOT provides a unified, statistically robust metric that accurately reflects the true Service Level Objectives experienced by end users.

Robust inference infrastructure is the bedrock of applied AI. By transitioning from constrained micro-benchmarking scripts to unbiased, distributed evaluation methodologies, the community can ensure that future architectural optimizations are driven by mathematically sound, reproducible data.

\section{\textbf{Acknowledgements}}

We are grateful to all the contributors of the Inference Perf project, especially Yuan Tang, Sachin Varghese, Brendan Slabe and Chen Wang for their vital role in helping develop and maintain it.

\medskip
{
\small
\bibliographystyle{unsrtnat} 
\bibliography{references}

@inproceedings{reddi2020mlperf,
  title={MLPerf Inference Benchmark},
  author={Reddi, V. J. and others},
  booktitle={2020 ACM/IEEE 47th Annual International Symposium on Computer Architecture (ISCA)},
  pages={446--459},
  year={2020},
  organization={IEEE},
  doi={10.1109/ISCA45697.2020.00045}
}

@article{kwon2023vllm,
  title={vLLM: Easy, Fast, and Cheap LLM Serving with PagedAttention},
  author={Kwon, Woosuk and Li, Zhuohan and Zhuang, Siyuan and Sheng, Ying and Zheng, Lianmin and Yu, Cody Hao and Rostaing, Joseph and Zhang, Hao and Stoica, Ion},
  journal={arXiv preprint arXiv:2309.06180},
  year={2023}
}

@article{zheng2023sglang,
  title={SGLang: Efficient Execution of Structured Language Model Programs},
  author={Zheng, Lianmin and Li, Li and Zhang, Hao and Zhuang, Yonghao and Chen, Zhijie and Huang, Yanping and Morris, Meredith Ringel and Gonzalez, Joseph E. and Stoica, Ion},
  journal={arXiv preprint arXiv:2312.07104},
  year={2023}
}

@misc{hf2023tgi,
  title={Text Generation Inference},
  author={{Hugging Face}},
  year={2023},
  howpublished={GitHub repository},
  url={https://github.com/huggingface/text-generation-inference}
}

@misc{grafana2021k6,
  title={k6: Open-source load testing tool},
  author={{Grafana Labs}},
  year={2021},
  url={https://k6.io/}
}

@misc{heyman2011locust,
  title={Locust: An open source load testing tool},
  author={Heyman, Jonatan and Bystr{\"o}m, Carl and Hamr{\'e}n, Joakim and Heyman, Hugo and Holmberg, Lars},
  year={2011},
  url={https://locust.io/}
}

@misc{vllm2023benchmarks,
  title={vLLM Benchmarks},
  author={{vLLM Team}},
  year={2023},
  howpublished={GitHub repository},
  url={https://github.com/vllm-project/vllm/tree/main/benchmarks}
}

@misc{nvidia2024genaiperf,
  title={GenAI-Perf. Part of Triton Inference Server},
  author={{NVIDIA}},
  year={2024},
  howpublished={GitHub repository},
  url={https://github.com/triton-inference-server/perf_analyzer/tree/main/genai-perf}
}

@misc{semianalysis2025inferencex,
  title={Inference X},
  author={{Semi Analysis}},
  year={2025},
  url={https://inferencex.semianalysis.com/}
}

@misc{semianalysis2025comp,
  title={Inference X Competitive Benchmarks},
  author={{Semi Analysis}},
  year={2025},
  url={https://newsletter.semianalysis.com/p/inferencemax-open-source-inference}
}

@article{williams2009roofline,
  title={Roofline: an insightful visual performance model for multicore architectures},
  author={Williams, Samuel and Waterman, Andrew and Patterson, David},
  journal={Communications of the ACM},
  volume={52},
  number={4},
  pages={65--76},
  year={2009},
  doi={10.1145/1498765.1498785}
}

@misc{chen2022transformer,
  title={Transformer Inference Arithmetic},
  author={Chen, Carol},
  year={2022},
  url={https://kipp.ly/blog/transformer-inference-arithmetic/}
}

@article{shoeybi2019megatron,
  title={Megatron-lm: Training multi-billion parameter language models using model parallelism},
  author={Shoeybi, M. and Patwary, M. and Puri, R. and LeGresley, P. and Casper, J. and Catanzaro, B.},
  journal={arXiv preprint arXiv:1909.08053},
  year={2019}
}

@article{narayanan2021efficient,
  title={Efficient large-scale language model training on GPU clusters using Megatron-LM},
  author={Narayanan, Deepak and Shoeybi, Mohammad and Casper, Jared and LeGresley, Patrick and Patwary, Mostofa and Korthikanti, Vijay Anand and Vainbrand, Dmitri and Kashinkunti, Prethvi and Bernauer, Julie and Catanzaro, Bryan and others},
  journal={arXiv preprint arXiv:2104.04473},
  year={2021},
  url={https://arxiv.org/abs/2104.04473}
}

@article{li2020pytorch,
  title={PyTorch Distributed: Experiences on accelerating data parallel training},
  author={Li, Shen and Zhao, Yanli and Varma, Rohan and Salpekar, Omkar and Noordhuis, Pieter and Li, Teng and Paszke, Adam and Smith, Jeff and Vaughan, Brian and Damania, Pritam and Chintala, Soumith},
  journal={arXiv preprint arXiv:2006.15704},
  year={2020},
  url={https://arxiv.org/abs/2006.15704}
}

@article{shazeer2017outrageously,
  title={Outrageously large neural networks: The sparsely-gated mixture-of-experts layer},
  author={Shazeer, N. and Mirhoseini, A. and Maziarz, K. and Davis, A. and Le, Q. and Hinton, G. and Dean, J.},
  journal={arXiv preprint arXiv:1701.06538},
  year={2017}
}

@article{lepikhin2020gshard,
  title={GShard: Scaling giant models with conditional computation and automatic sharding},
  author={Lepikhin, D. and Lee, H. and Xu, Y. and Chen, D. and Firat, O. and Huang, Y. and Krikun, M. and Shazeer, N. and Chen, Z.},
  journal={arXiv preprint arXiv:2006.16668},
  year={2020}
}

@misc{mlcommons2024v4,
  title={MLPerf v4.0 LLM Benchmarks},
  author={{MLCommons}},
  year={2024},
  url={https://mlcommons.org/2024/03/mlperf-inference-v4/}
}

@misc{nvidia2023tensorrtllm,
  title={TensorRT-LLM (Version 1.0)},
  author={{NVIDIA}},
  year={2023},
  howpublished={GitHub repository},
  url={https://github.com/NVIDIA/TensorRT-LLM}
}

@misc{artificialanalysis2025,
  title={Artificial Analysis Providers Leaderboard},
  author={{Artificial Analysis}},
  year={2025},
  url={https://artificialanalysis.ai/leaderboards/providers}
}

@inproceedings{hu2022lora,
  title={LoRA: Low-Rank Adaptation of Large Language Models},
  author={Hu, E. J. and Shen, Y. and Wallis, P. and Allen-Zhu, Z. and Li, Y. and Wang, S. and Wang, L. and Chen, W.},
  booktitle={International Conference on Learning Representations (ICLR)},
  year={2022}
}

@inproceedings{zhong2024distserve,
  title={DistServe: Disaggregating Prefill and Decoding for Goodput-Optimized Large Language Model Serving},
  author={Zhong, Y. and Liu, S. and Chen, J. and Hu, J. and Zhu, Y. and Liu, X. and Jin, X. and Zhang, H.},
  booktitle={18th USENIX Symposium on Operating Systems Design and Implementation (OSDI 24)},
  pages={193--210},
  year={2024}
}

@inproceedings{patel2024splitwise,
  title={Splitwise: Efficient generative LLM inference using phase splitting},
  author={Patel, P. and Choukse, E. and Zhang, C. and Shah, A. and Goiri, {\'I}. and Maleki, S. and Bianchini, R.},
  booktitle={Proceedings of the 51st Annual International Symposium on Computer Architecture (ISCA)},
  pages={118--132},
  year={2024}
}

@article{jaiswal2025sageserve,
  title={SageServe: Optimizing LLM Serving on Cloud Data Centers with Forecast Aware Auto-Scaling},
  author={Jaiswal, S. and Jain, K. and Simmhan, Y. and Parayil, A. and Mallick, A. and Wang, R. and Amant, R. S. and Bansal, C. and Ruhle, V. and Kulkarni, A. and Kofsky, S.},
  journal={Proceedings of the ACM on Measurement and Analysis of Computing Systems},
  volume={9},
  number={3},
  pages={1--24},
  year={2025}
}

@misc{rayproject2025llmperf,
  title={LLM Perf: A Tool for the Performance Evaluation of LLM APIs},
  author={{Ray Project}},
  year={2025},
  howpublished={GitHub repository},
  url={https://github.com/ray-project/llmperf}
}

@misc{neuralmagic2024guidellm,
  title={GuideLLM: Scalable Inference and Optimization for Large Language Models},
  author={{Neural Magic, Inc.}},
  year={2024},
  howpublished={GitHub repository},
  url={https://github.com/vllm-project/guidellm}
}

@misc{nvidia2024aiperf,
  title={AI Perf},
  author={{NVIDIA}},
  year={2024},
  howpublished={GitHub repository},
  url={https://github.com/ai-dynamo/aiperf}
}

@misc{sglangbenchmarks,
  title={{SGLang} Benchmarking Utilities},
  author={{SGL-Project Team}},
  year={2024},
  howpublished={GitHub repository},
  url={https://github.com/sgl-project/sglang/tree/main/benchmark}
}

@misc{llmd2026inference,
  title={LLM-D Inference Simulator},
  author={{LLM-D Team}},
  year={2026},
  howpublished={GitHub repository},
  url={https://github.com/llm-d/llm-d-inference-sim}
}

@book{kleinrock1975queueing,
  author    = {Kleinrock, Leonard},
  title     = {Queueing Systems, Volume 1: Theory},
  publisher = {Wiley-Interscience},
  address   = {New York, NY},
  year      = {1975},
  isbn      = {978-0471491101}
}

@misc{inferenceperf2026,
  title = {{Inference Perf}},
  author = {{Kubernetes SIGs}},
  year = {2026},
  howpublished = {GitHub repository},
  url = {https://github.com/kubernetes-sigs/inference-perf}
}
}

\newpage
\appendix

\section{Appendix}

\subsection{Latency profiles}

This section provides visual representations of latency profiles, including a general illustration showing the pre- and post- saturation regimes and the ideal operating zone for a model server (Figure 3) and specific results obtained using the Gemma-3-1b-it model on an 8xH100 GPU cluster (Figures 4, 5, 6).

\begin{figure}[H]
    \centering
    \includegraphics[width=0.5\linewidth]{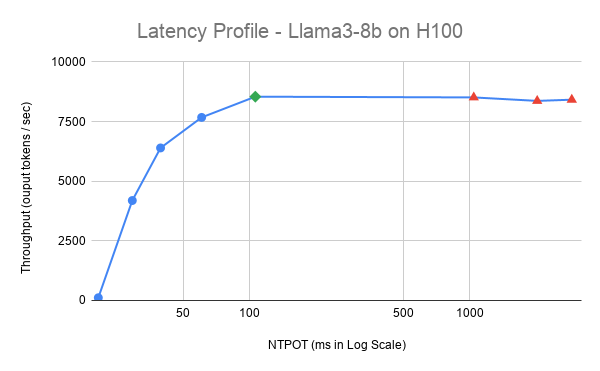}
    \caption{Latency profile illustration showing the ideal operating zone (blue dot), saturation point (green diamond) and post-saturation points (red triangle) where latency SLOs will be severely affected}
    \label{fig:3}
\end{figure}

\begin{figure}[H]
    \centering
    \includegraphics[width=1\linewidth]{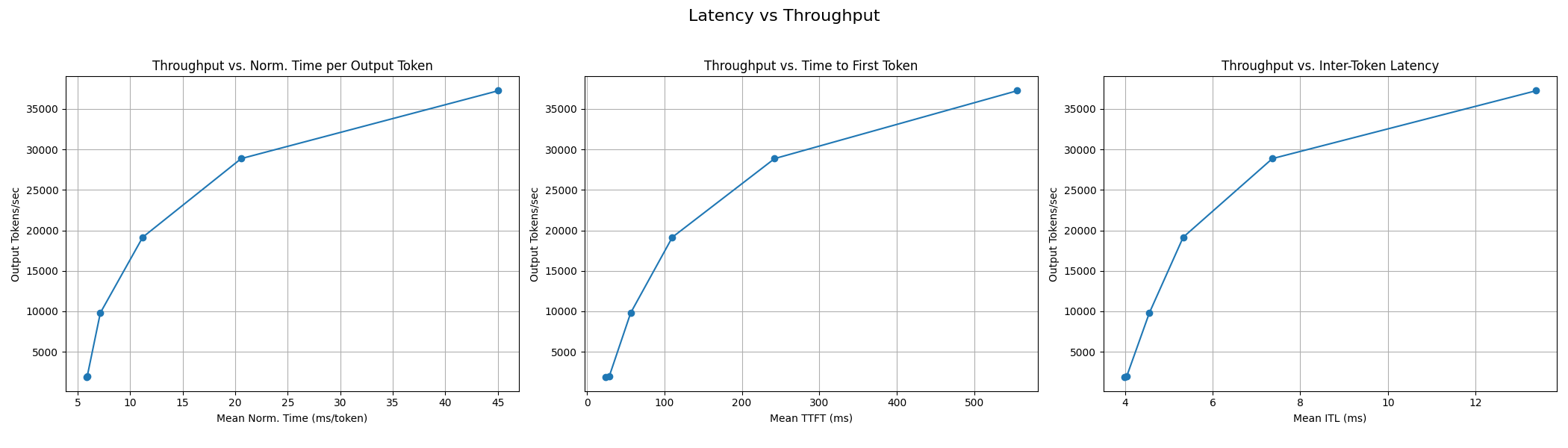}
    \caption{Latency profile showing the throughput vs latency curve for the Gemma-3-1b-it model on 8xH100 GPUs trending towards saturation}
    \label{fig:4}
\end{figure}

\begin{figure}[H]
    \centering
    \includegraphics[width=1\linewidth]{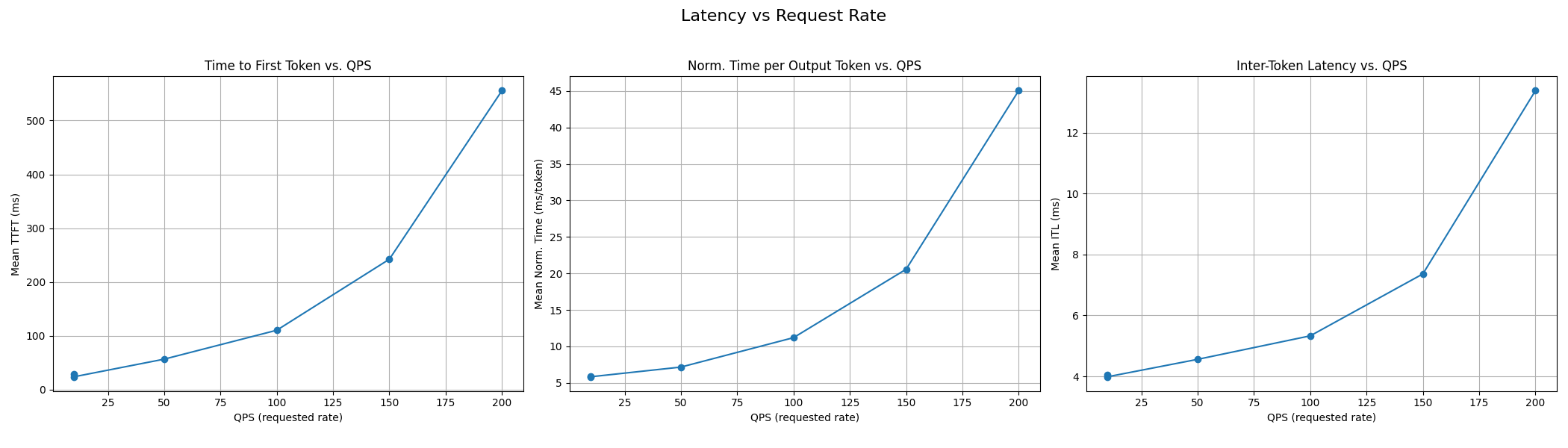}
    \caption{Latency vs QPS chart showing the latency growth relative to load and how NTPOT is able to provide a normalized view of latency}
    \label{fig:5}
\end{figure}

\begin{figure}[H]
    \centering
    \includegraphics[width=1\linewidth]{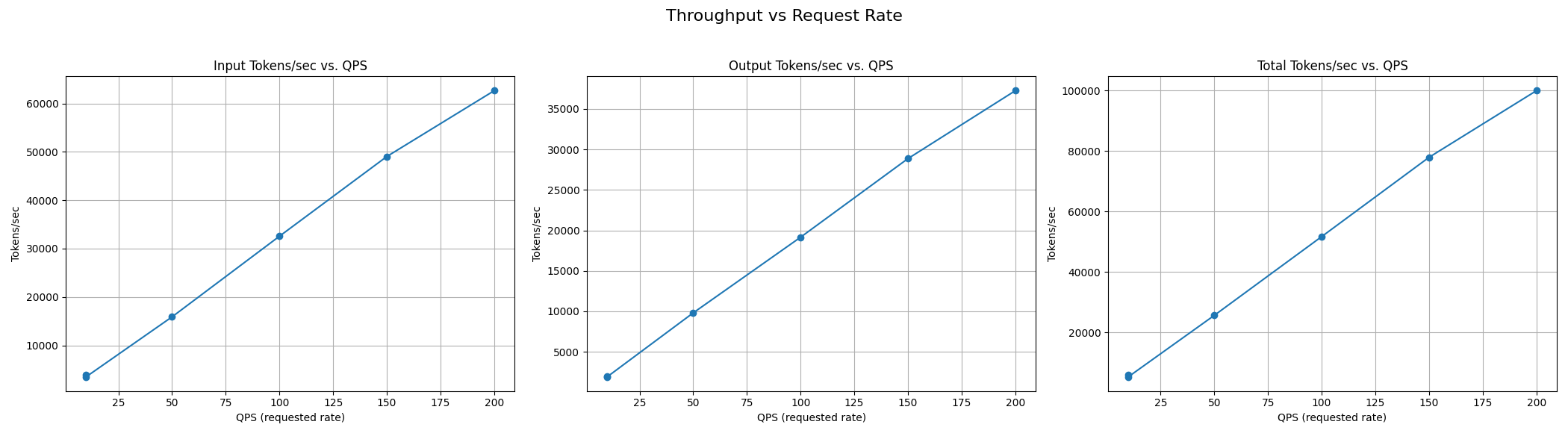}
    \caption{Throughput vs QPS chart showing the throughput growth relative to load and how even at 200 QPS we are not fully saturated, which shows that the Python GIL bottleneck of single-process tools can affect the outcome even at a relatively low QPS like 200 when combined with the result in Table 2}
    \label{fig:6}
\end{figure}

\subsection{Factors affecting benchmarking reproducibility}

This section explores the impact of experimental variables on benchmarking reproducibility, quantifying throughput variance due to probabilistic sampling configurations (temperature settings) and analyzing the effects of dataset processing discrepancies between different benchmarking tools. This shows the need for standardized configuration across tools in the ecosystem with enough specificity to reproduce them without discrepancies.

\begin{table}[H]
\centering
\caption{Variance in throughput induced by probabilistic sampling configurations}
\label{tab:3}
\begin{tabular}{ l  l }
\toprule
\textbf{Configuration} & \textbf{Output Token Throughput (tok/s)} \\
\midrule

Batch, 1000 requests (Server/Model Default, Temp=0.7) & 5573 \\

Batch, 1000 requests (Temp=0) & 6767 \\
\bottomrule

\end{tabular}

\end{table}

\begin{table}[H]
\centering
\caption{Effect of dataset processing by benchmarking tools}
\label{tab:4}
\begin{tabular}{ l  l  l }
\toprule
\textbf{Tool} & \textbf{Configuration} & \textbf{Input Tokens Processed} \\
\midrule

Tool 1 (Inference X)& ShareGPT, 3000 requests at 10 QPS & 702,412 \\

Tool 2 (Inference Perf)& ShareGPT, 3000 requests at 10 QPS & 1,052,100 \\
\bottomrule

\end{tabular}

\end{table}

\subsection{Multi-process request workflow in Inference Perf}

This section illustrates the multi-process request workflow implemented in the benchmarking framework, detailing how the architecture manages load partitioning to mitigate measurement bias.

\begin{figure}
    \centering
    \includegraphics[width=1\linewidth]{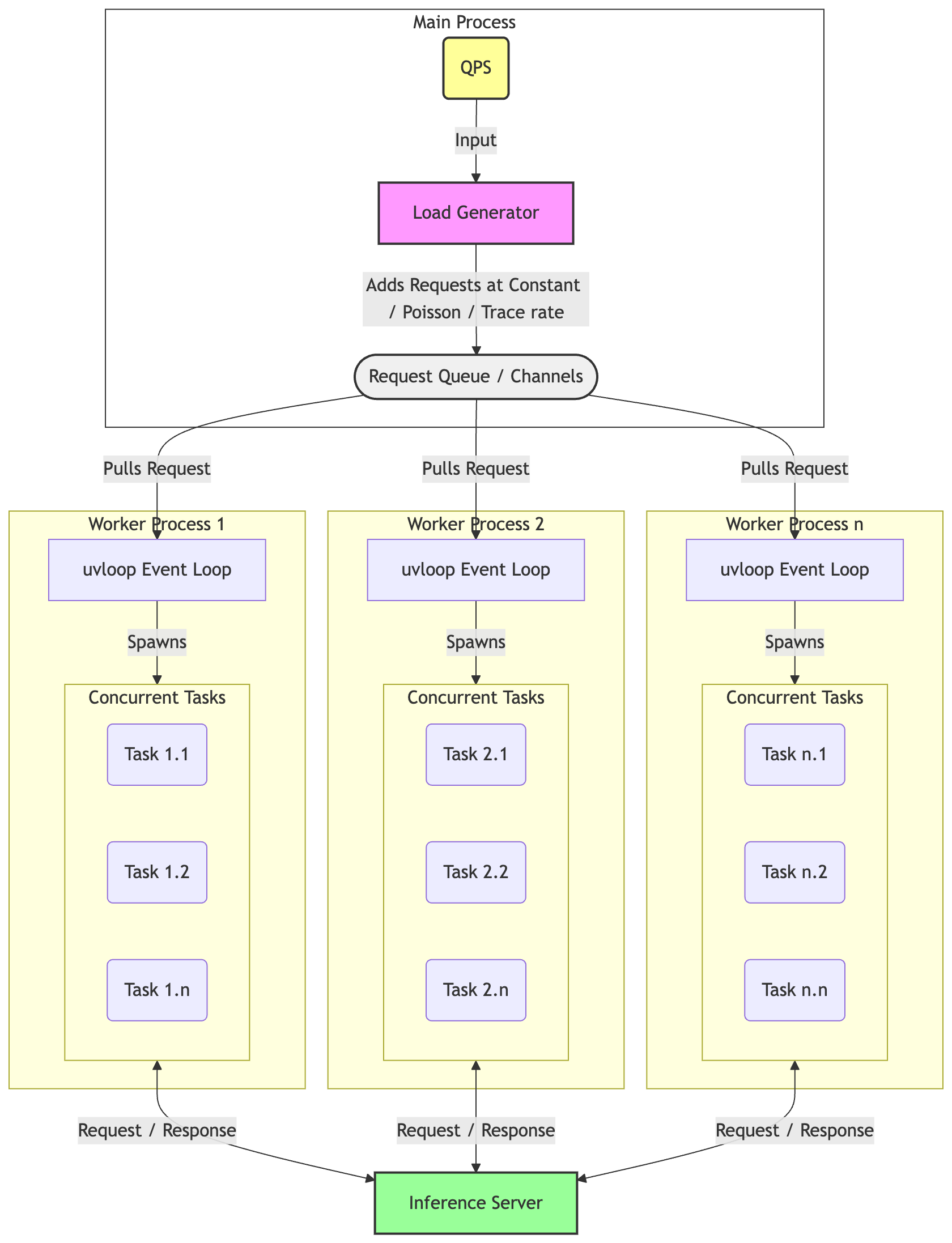}
    \caption{Multi-process request workflow in Inference Perf showing how the architecture overcomes the single event loop and CPU exhaustion issue in single-process benchmarking tools by using a main orchestration process and $n$ worker processes to evenly distribute the load}
    \label{fig:7}
\end{figure}


\end{document}